\documentclass[letterpaper, 10 pt, conference]{ieeeconf}  
\usepackage{graphicx}
\usepackage{amssymb}
\usepackage{amsopn}
\usepackage{amsmath}

\IEEEoverridecommandlockouts                              

\title{\LARGE \bf
HyperDCM: Dynamic Cluster Memory Replay in Hyperbolic Space \\ for Continual Robotic Navigation Across Scenes
}

\author{Zhengfei Lu$^{1}$, Jian Yang$^{2}$$^{\dagger}$, Muyu Wang$^{1}$,
Shaowen Chen$^{1}$, Jinpeng Mi$^{3}$, Ke Li$^{2}$, Xiong You$^{2}$, \\
Qi Wu$^{4}$, Xuan Tang$^{1}$, Xian Wei$^{1}$
\thanks{
This research is supported by the National Natural Science Foundation of China (No. 42130112), the Ministry of Industry and Information Technology of China, the National Science and Technology Major Project (Nos.2025ZD1606804 and 2025ZD1601304), the General Program of Shanghai Natural Science Foundation (No. 24ZR1419800), and the Shanghai Frontiers Science Center of Molecule Intelligent Syntheses.
}
\thanks{$^{1}$Software Engineering Institute, East China Normal University, China; 
$^{2}$School of Geospatial Information, Information Engineering University, China; 
$^{3}$University of Shanghai for Science and Technology, China; 
$^{4}$Shanghai Jiao Tong University, China.
}
\thanks{$^{\dagger}$Corresponding author: {\tt\small jian.yang@tum.de}.}
}

\begin{document}

\maketitle
\thispagestyle{empty}
\pagestyle{empty}

\begin{abstract}

Continual learning in visual navigation remains challenging due to catastrophic forgetting and the difficulties associated with adapting to diverse and evolving environments. 
To address these issues, we propose Hyperbolic Dynamic Cluster Memory (HyperDCM), a structure-aware memory mechanism that enhances diffusion policy-based navigation through scene graph modeling and principled memory replay. HyperDCM extracts semantic scene triples from RGB observations using large vision–language models, encodes them into scene graph embeddings via a Relational Graph Convolutional Network (R-GCN), and projects the embeddings into hyperbolic space to enhance structural separability and retention in continual navigation.
A dynamic clustering and structure-sensitive update strategy selects representative samples for memory replay, thereby preserving knowledge diversity and mitigating catastrophic forgetting. Experiments on multi-scene indoor and outdoor datasets demonstrate that HyperDCM achieves superior retention of past navigation capabilities and improved generalization compared to representative continual learning baselines adapted to diffusion policy navigation.
\end{abstract}

\begin{figure*}
    \centering
    \includegraphics[width=1\linewidth]{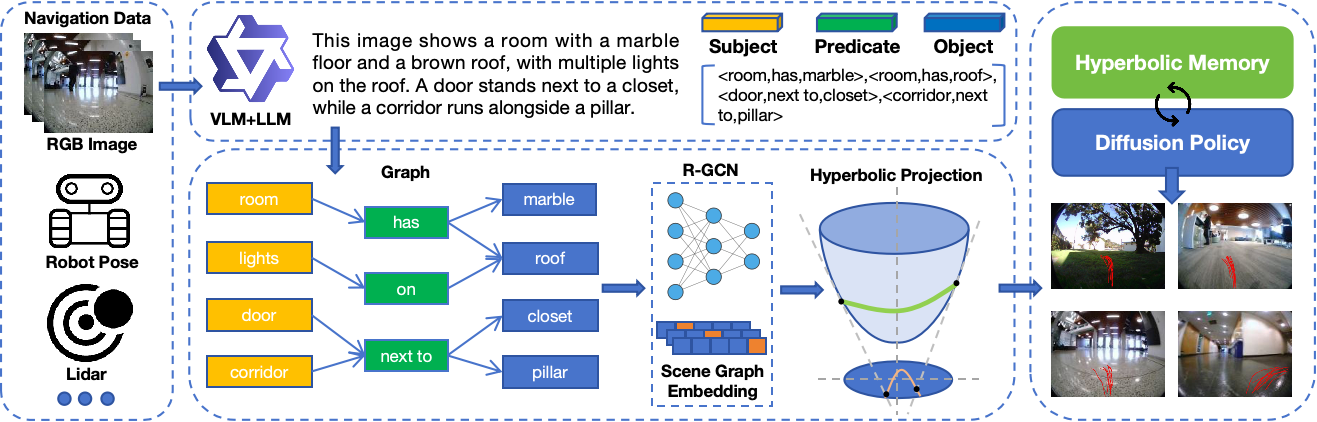}
    \caption{\textbf{Overview of the proposed HyperDCM mechanism.} 
Navigation data from robot trajectories is collected, where RGB observations are processed by vision–language models to generate semantic descriptions, which are parsed into (Subject, Predicate, Object) triples. These triples are used to construct scene graphs, encoded via an R-GCN, and projected into hyperbolic space to capture hierarchical relations. During diffusion policy training, hyperbolic-space dynamic memory replay preserves diverse samples and mitigates catastrophic forgetting.}
    \label{fig:scene_extract}
\end{figure*}

\section{Introduction}

Recently, visual navigation has attracted increasing research attention~\cite{richter2017safe, gupta2017cognitive, mayo2021visual, shah2023vint, liu2024ok, zhang2025dual}, due to its wide application in indoor and outdoor navigation.
However, real-world environments are often dynamic and complex. 
Robotic navigation must continually operate in diverse and evolving conditions.
It relies on a reliable perception of environmental cues and adaptive actions to reach targets.
Traditional policy learning methods, such as regression-based or discriminative approaches, struggle to capture the multi-modal nature of navigation tasks. 
As a result, their generalization ability in complex environments remains limited.  

Against this backdrop, diffusion policy has emerged as a promising generative learning paradigm~\cite{chi2023diffusion}. 
Such models can represent multi-modal action distributions and refine noisy actions into feasible trajectories. 
In navigation tasks, diffusion policy-based approaches generate sequences of future actions over multiple timesteps, offering stronger foresight and improved generalization~\cite{sridhar2024nomad}.  

Nevertheless, most robot diffusion policy methods are designed for a single situation and trained on static and closed datasets. 
This significantly restricts adaptability when deployed in open-world environments.
In practical applications, robots must handle new scenes, novel task categories, dynamic obstacles, and heterogeneous platforms~\cite{kim2025open, ye2024online, dai2023domain, liu2023ai, chen2025target, wang2024navformer}. 
These challenges widen the gap between offline training and real-world deployment. 
Offline-trained models are also difficult to update, while repeated retraining is costly, prone to overfitting, and often discards prior knowledge.
Current policy models are vulnerable to catastrophic forgetting under continual adaptation~\cite{van2020brain}.
This weakness is particularly noticeable in diffusion policies because their action generation heavily relies on pixel-level or vectorized features, without explicit memory and structural awareness. This limitation prevents these methods from retaining relational diversity across experiences, which degrades performance in continually changing environments. 

To overcome this issue, a representation capable of capturing both object relations and hierarchical structures is needed. 
Pixel-level or vectorized features can only answer ``what is in the scene'' but not ``how objects are related.'' 
Scene graphs provide such a representation by encoding observations as subject--predicate--object triplets~\cite{li2017scene}, delivering richer semantic information for continual learning. 
Furthermore, hyperbolic geometry excels at modeling hierarchical and tree-like structures. 
It offers low-distortion embeddings for scene graphs~\cite{peng2021hyperbolic}, while Euclidean embeddings may struggle to preserve such relations.
By combining scene graphs with hyperbolic embeddings, structure-aware memory replay can be built, which improves adaptability and reduces forgetting in open-world navigation.
Based on these insights, we propose Hyperbolic Dynamic Cluster Memory (HyperDCM). 
It is a structure-aware memory replay mechanism designed to enhance the adaptability and anti-forgetting ability of diffusion policies in open-world environments. 
The main contributions of this work are summarized as follows:  
\begin{enumerate}
    \item We propose a structure-aware memory representation by encoding scene graphs into hyperbolic space, preserving hierarchical and relational information beyond Euclidean embeddings.  
    \item We introduce a dynamic clustering and structure-sensitive update strategy, which organizes and updates samples using hyperbolic distances and centroid replacement. This strategy maintains knowledge diversity and mitigates forgetting in multi-scene continual training.  
    \item We demonstrate through experiments on multi-scene indoor and outdoor datasets that the proposed HyperDCM retains knowledge more effectively and achieves better generalization than NoMaD under continual learning settings.
\end{enumerate}

\section{Related Work}

\subsection{Diffusion Policy for Navigation}
Diffusion models have achieved remarkable success in image generation tasks~\cite{ho2020denoising, croitoru2023diffusion}, and their potential is now being explored in robotics. 
Unlike traditional regression methods, diffusion policy treats action generation as a process of conditional diffusion followed by denoising sampling, enabling effective modeling of high-dimensional continuous actions~\cite{chi2023diffusion}.  

Diffusion-based policies have also been applied to navigation tasks. 
A goal-conditioned framework enables flexible switching between navigation and exploration~\cite{sridhar2024nomad}. 
Subsequent studies integrate reinforcement learning, transformer architectures, and cost-guided mechanisms to improve policy generation and goal alignment~\cite{zhang2024versatile,zeng2025navidiffusor}. 
More recent work extends diffusion policies to graph-conditioned planning~\cite{cao2025dare, huang2025kiterunner} and dynamic obstacle handling~\cite{yu2024ldp}.

These studies collectively highlight the versatility of diffusion policies across exploration, collision avoidance, crowd navigation, vision-language navigation, mapless global planning, and locomotion. 
Most existing works, however, are still trained on static datasets and focus on single-task or joint training settings, while continual adaptation across multiple navigation scenes has received little attention.

\subsection{Continual Learning}

Continual learning (CL) studies how to enable models to adapt to sequential tasks without catastrophic forgetting~\cite{wang2024comprehensive}. 
Existing approaches are commonly categorized into three paradigms: 
regularization-based methods that constrain parameter updates~\cite{kirkpatrick2017overcoming,zenke2017continual,li2017learning}, 
replay-based strategies that store representative past samples and interleave them with current training~\cite{rebuffi2017icarl,chaudhry2019tiny}, 
and parameter-isolation techniques that allocate disjoint parameter subsets for different tasks~\cite{mallya2018packnet,serra2018overcoming}.  

Among these paradigms, replay-based approaches are particularly effective in practice. 
By maintaining a memory buffer $M$, replay methods jointly optimize current data and stored samples, substantially mitigating forgetting even under limited memory~\cite{chaudhry2019tiny}. 
Recent advances extend replay through generative models~\cite{chen2024stable} or parameter-efficient updates such as adapters and LoRA~\cite{smith2024continual,smith2023continual}. 
However, these approaches often overlook the structured nature of robotics and navigation, where experiences are inherently relational and hierarchical. This gap motivates our structure-aware replay paradigm, which leverages scene graphs and hyperbolic embeddings to preserve relational diversity in memory.  

\subsection{Scene Graph Embedding}

Scene graphs represent images as structured subject--predicate--object triplets, where nodes denote entities and edges encode semantic or spatial relations~\cite{khandelwal2024adaptive}. 
Compared to flat object detections, scene graphs capture relational dependencies and structural organization within a scene.

Graph-based reasoning models, such as graph convolutional networks~\cite{yang2018graph}, have demonstrated strong capability in modeling object interactions. 
However, most existing work focuses on visual understanding tasks, while the use of scene graph representations for continual robot navigation remains underexplored. 
In this work, we leverage scene graphs as structured embeddings to preserve relational diversity in memory.

\subsection{Hyperbolic Embedding}

Hyperbolic space, characterized by constant negative curvature, is well-suited for modeling hierarchical and tree-like structures. 
Compared with Euclidean space, it enables low-distortion embeddings in low dimensions~\cite{ganea2018hyperbolic,peng2021hyperbolic}. 

\begin{figure}
    \centering
    \includegraphics[width=1\linewidth]{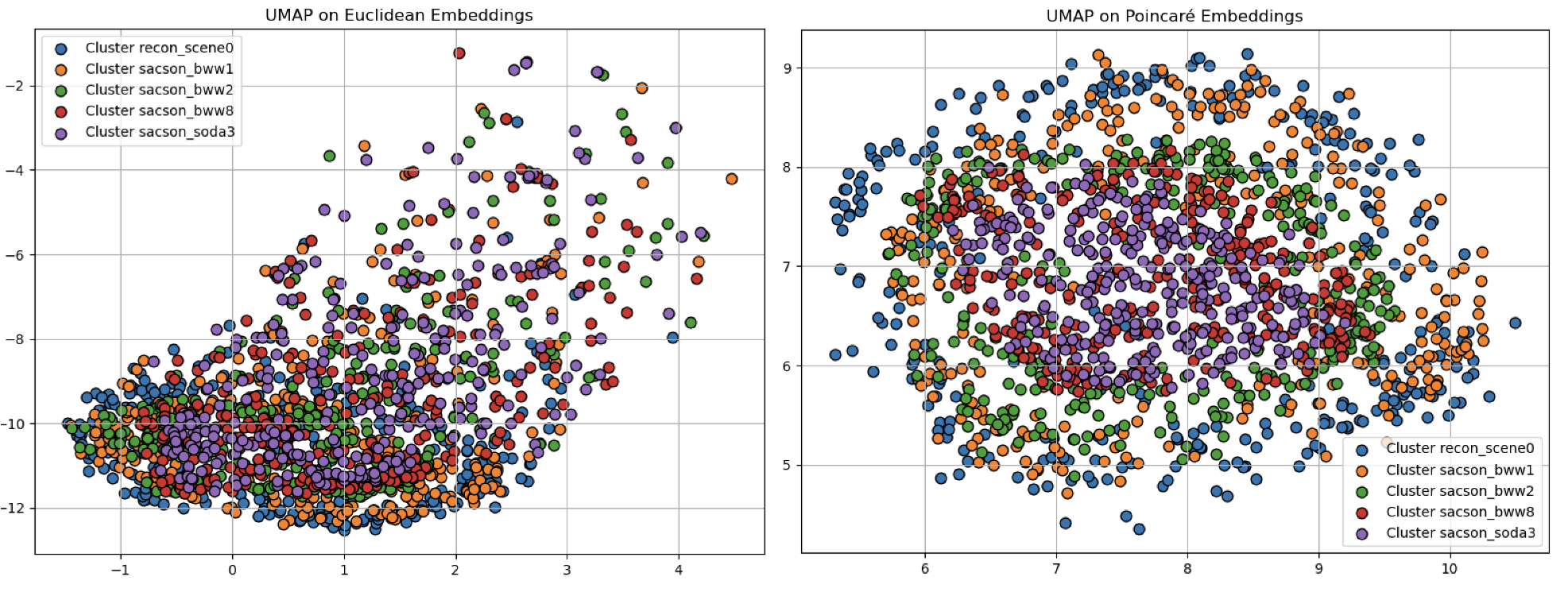}
    \caption{UMAP visualization comparing the Euclidean and Poincaré embeddings of scene-level memory samples. The hyperbolic space (right) shows better separation and local clustering, reflecting superior structural awareness in complex visual navigation settings.}
    \label{fig:embeeding_compare}
\end{figure}

Scene graphs inherently exhibit hierarchical semantics, which are difficult to faithfully preserve in Euclidean geometry. 
The Poincaré disk model has been shown to effectively capture hierarchical relations through geodesic distance~\cite{nickel2017poincare}.

Motivated by these properties, we project scene graph embeddings into Poincaré hyperbolic space and incorporate them into a dynamic memory mechanism for continual navigation.

\begin{figure*}
    \centering
    \includegraphics[width=1\linewidth]{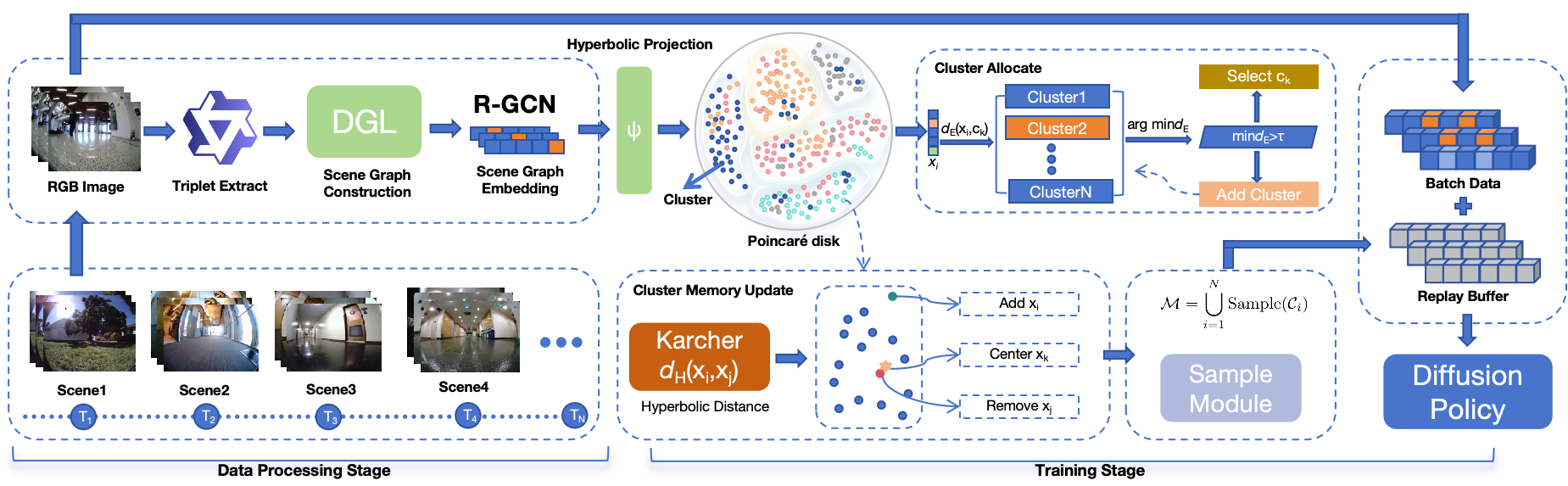}
    \vspace{-15pt}
    \caption{Dynamic cluster memory management in hyperbolic space. Each new sample computes its Euclidean distance to all cluster centroids, being assigned to the nearest cluster or initiating a new one if the distance exceeds a threshold and the cluster limit is not reached. Hyperbolic distance is then used to update more representative samples, with cluster centroids refined via the Karcher mean. During replay, a fixed number of samples are randomly drawn from multiple clusters and combined with the current batch to train the Diffusion Policy, preserving memory diversity and mitigating catastrophic forgetting.}
    \label{fig:framework_v01}
\end{figure*}

\section{The Proposed HyperDCM}

The proposed HyperDCM mechanism enhances continual diffusion-based visual navigation by integrating structured scene understanding with a hyperbolic memory representation. As shown in Fig.~\ref{fig:scene_extract}, the approach consists of three main stages: (1) structured scene graph representation, (2) hyperbolic embedding via Poincaré mapping, and (3) dynamic memory scheduling and updating.

\subsection{Structured Scene Graph Representation}
We first construct structured scene graph representations from raw RGB observations to capture semantic relationships among objects. 
Such structured modeling facilitates relational reasoning and improves generalization across multiple scenes in continual navigation tasks.

We utilize Qwen-VL to extract textual descriptions from images and employ Qwen3-7B to generate semantic triples $(s,r,o)$. 
These large models provide reliable relational extraction without handcrafted rule design.

To embed the scene graph, we use a Relational Graph Convolutional Network (R-GCN). R-GCN effectively models multiple types of edges and captures rich semantic information while preserving the topological structure of the graph. The resulting graph embedding \(z_i \in \mathbb{R}^{128}\) is used as the structured representation for subsequent hyperbolic mapping and memory clustering.

The R-GCN computes node embeddings using:
\[
h_i^{(l+1)} = \sigma\left( \sum_{r \in \mathcal{R}} \sum_{j \in \mathcal{N}_r(i)} \frac{1}{c_{i, r}} W_r^{(l)} h_j^{(l)} + W_0^{(l)} h_i^{(l)} \right),
\]
where $\mathcal{R}$ denotes relation types, $\mathcal{N}_r(i)$ represents neighbors of node $i$ under relation $r$, $c_{i,r}$ is a normalization constant, and $\sigma$ is an activation function. 
The weights $W_r$ and $W_0$ follow the standard R-GCN formulation and are fixed after preprocessing, as the R-GCN is used solely for offline embedding generation.

\subsection{Hyperbolic Embedding via Poincaré Mapping}
Hyperbolic space is characterized by constant negative curvature, enabling efficient representation of hierarchical structures. 
In the Poincaré disk model, geodesic distances grow rapidly near the boundary, allowing embeddings to allocate greater representational capacity to lower-level nodes while preserving their hierarchical relations.

In our setting, scene graphs often exhibit tree-like semantics (e.g., “room contains table, table supports object”), 
where certain nodes branch into multiple relations. 
Euclidean embeddings struggle to preserve such branching, leading to distortions in similarity modeling. 
By contrast, mapping graph embeddings into hyperbolic space ensures that structurally distinct samples are naturally placed further apart, 
resulting in clearer cluster separation as shown in Fig.~\ref{fig:embeeding_compare}. 
This separation provides a stronger structural prior, enabling memory replay to retain diverse and informative samples across scenes better.

We adopt the Poincaré disk model for hyperbolic embedding due to its suitability for hierarchical representation and closed-form distance computation. 
Samples with richer relational structures are naturally positioned farther from the origin, improving structural separability.

The scene graph embedding $z_i$ is projected into a hyperbolic space using the Poincaré disk model:
\[
\phi(z_i) = \frac{z_i}{\sqrt{c}(\|z_i\| + \epsilon)} \cdot \tanh(\sqrt{c} \|z_i\|),
\]
where $c$ is the curvature parameter and $\epsilon$ is a small constant for numerical stability. We use a fixed curvature $c$ across all scenes, determined beforehand and kept unchanged during final evaluation.

\subsection{Dynamic Memory Scheduling and Update}
Having obtained hyperbolic scene embeddings, we next describe how samples are clustered and scheduled for replay to mitigate forgetting. The overall procedure is illustrated in Fig.~\ref{fig:framework_v01}. 
We divide memory into $C$ clusters, each with capacity $M$. Here, $C$ denotes the number of memory clusters and $M$ is the per-cluster capacity; 
$d_E(\cdot,\cdot)$ and $d_{\mathbb{H}}(\cdot,\cdot)$ denote Euclidean and hyperbolic geodesic distances, respectively. For each new scene embedding $\phi(z_i)$:

Compute the Euclidean distance $d_E$ to the center of all clusters $\{c_k\}_{k=1}^C$:
    \[
    \min_{k \in \{1, ..., C\}} d_E(\phi(z_i), c_k).
    \]

If the minimum distance exceeds a threshold $\tau$ and the cluster capacity permits, a new cluster is created; otherwise, the sample is added to the nearest cluster. We use a fixed clustering threshold $\tau$ across all scenes and keep it unchanged during final evaluation. A maximum cluster cap is used, after which new samples are assigned to the nearest cluster to bound memory growth. We use Euclidean distance only for fast cluster assignment, while hyperbolic distance is employed for subsequent updates to respect manifold geometry.

To robustly represent the structural center of each memory cluster in hyperbolic space, we employ the Karcher mean, a generalization of the Euclidean mean to Riemannian manifolds. Unlike the simple arithmetic mean, the Karcher mean respects the underlying geometry of the Poincaré disk model, ensuring that the computed center remains within the hyperbolic manifold and better reflects geodesic distances between points. For each cluster, the structural center is maintained using the Karcher mean:
    \[
    \mu = \arg\min_{y \in \mathcal{M}} \sum_{i=1}^{N} d_{\mathbb{H}}^2(y, x_i),
    \]    
where $\mu$ is the Karcher mean. $\mathcal{M}$ denotes the hyperbolic space memory, and $d_{\mathbb{H}}(y, x_i)$ represents the geodesic distance between the candidate center $y$ and each sample point $x_i$. In practice, the Karcher mean is updated with a small fixed number of Riemannian gradient steps within the affected cluster. Since both the number of clusters and the cluster capacity are bounded, this update introduces only bounded overhead and does not require retraining the diffusion policy.

The Poincaré distance between two points $\mathbf{u}$, $\mathbf{v} \in \mathcal{M}^n$ is defined as:
\[
d_{\mathbb{H}}(\mathbf{u}, \mathbf{v}) = \operatorname{arcosh} \left( 1 + \frac{2 \| \mathbf{u} - \mathbf{v} \|^2}{(1 - \| \mathbf{u} \|^2)(1 - \| \mathbf{v} \|^2)} \right),
\]
where $\|\cdot\|$ denotes the Euclidean norm. This formulation captures the curved geometry of hyperbolic space and enables structurally faithful distance calculations, which are crucial for tasks involving hierarchical or graph-structured data.

To maintain the structural richness within a cluster, we propose a structure-sensitive update strategy as follows:

Let $\bar{x}_k$ be the Karcher mean (centroid) of cluster $C_k$, defined by:
\[
\bar{x}_k = \arg\min_{x \in \mathbb{D}^n} \sum_{x_j \in C_k} d_{\mathbb{H}}^2(x, x_j).
\]

\begin{figure*}
    \centering
    \includegraphics[width=1\linewidth]{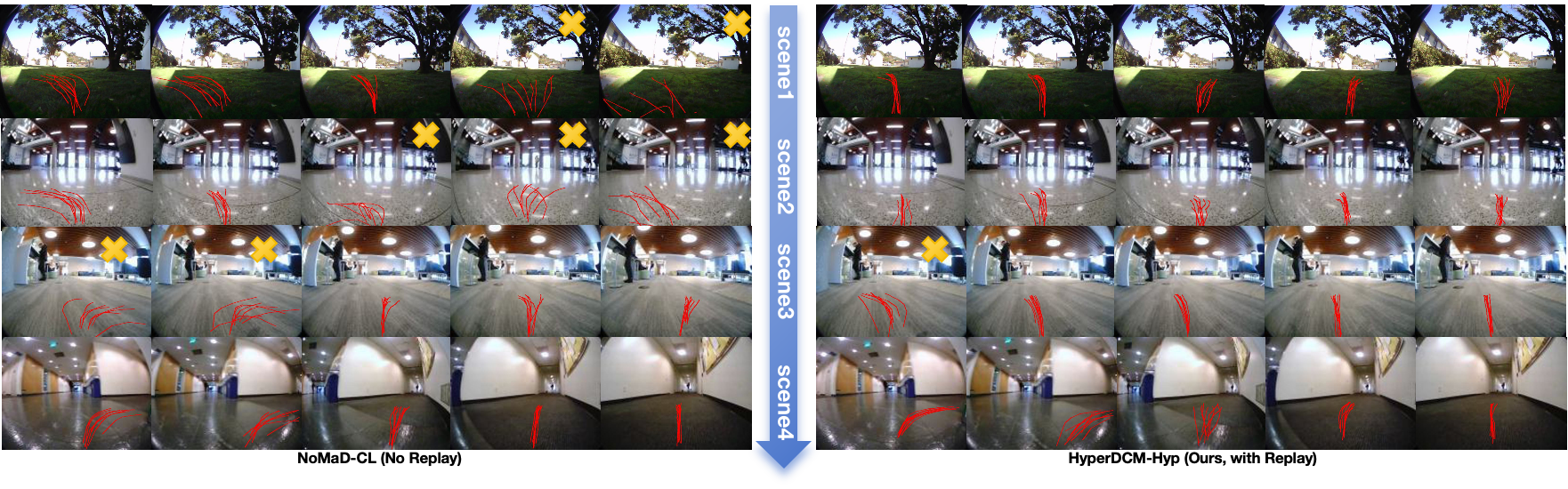}
    \vspace{-25pt}
    \caption{Qualitative comparison of predicted waypoint trajectories across four representative scenes. 
Left: continual NoMaD baseline. Right: HyperDCM-Hyp. 
Earlier scenes exhibit trajectory drift under NoMaD, while HyperDCM maintains stable predictions.}
    \label{fig:eval_show}
\end{figure*}

Then compute the hyperbolic distance $d_{\mathbb{H}}(\tilde{x}_i, \bar{x}_k)$ from a new sample $\tilde{x}_i$ to the centroid $\bar{x}_k$. Identify the current sample $x_j$ in cluster $C_k$ that is closest to $\bar{x}_k$ in terms of hyperbolic distance. If $d_{\mathbb{H}}(\tilde{x}_i, \bar{x}_k) > d_{\mathbb{H}}(x_j, \bar{x}_k)$, we consider $\tilde{x}_i$ structurally richer and replace $x_j$ with $\tilde{x}_i$. The above strategy can be formalized as:
If $d_{\mathbb{H}}(\tilde{x}_i,\bar{x}_k) > \min_{x_j \in C_k} d_{\mathbb{H}}(x_j,\bar{x}_k)$, then replace $x_j$ with $\tilde{x}_i$.
\subsection{Memory Replay Strategy}
To enhance diversity during replay, we uniformly sample n instances from the $C$ memory clusters:
\[
\mathcal{B} = \text{Sample}(\cup_{k=1}^{C} \text{Cluster}_k, \text{size}=n).
\]

These samples are used alongside current data to finetune the diffusion policy, preserving past knowledge while learning new scenes.

\section{EXPERIMENTS}

\subsection{Training and Implementation Details}
We train the diffusion policy model on an NVIDIA RTX 4090 GPU.
As our baseline, we adopt NoMaD, which demonstrates strong performance in diffusion-based navigation.
The original NoMaD follows a joint training paradigm, where all scenes are trained simultaneously.
To evaluate continual learning performance, we reorganize the dataset into multiple scene splits and train them sequentially.
Each scene is trained for 20 epochs, and after each stage we save a checkpoint to evaluate retention on previously encountered scenes.

\subsection{Datasets}

For indoor evaluation, we adopt the SACSoN dataset~\cite{hirose2023sacson}, which provides multi-scene indoor navigation trajectories. For outdoor evaluation, we use the Recon dataset~\cite{shah2021rapid}, which contains real-world robot trajectories across diverse outdoor environments. We construct five distinct scene splits from these datasets. For each scene, $60\%$ of the trajectory data is used for training and the remaining $40\%$ for testing. Each trajectory contains robot-view images, pose information, and goal-conditioned annotations, which are required for diffusion-policy training.

\subsection{Evaluation Metrics and Protocol}
We first evaluate under the trajectory prediction setting following NoMaD. To evaluate performance under continual learning, we primarily use Success Rate (SR):
\[
\text{SR} = \frac{1}{N} \sum_{i=1}^{N} \mathbf{1}\!\left[d(s_i, g_i) \leq \delta \right],
\]
where $N$ is the total number of trajectories, $s_i$ is the final state of trajectory $i$, $g_i$ is the goal, 
$d(\cdot)$ is the distance function, and $\delta$ is the success threshold.
For clarity, we denote \texttt{1$\sim$X} as the model trained sequentially from Scene 1 up to Scene X. 
Evaluation is conducted using the checkpoint obtained at the end of the corresponding training sequence. 

For each test trajectory, we use the next image in the sequence as the navigation goal and generate a trajectory 
from the current observation using the trained diffusion policy. If over 90\% of predicted waypoints in the 
trajectory align with the ground truth, we count it as a successful navigation case.  

We additionally report the Drop metric, defined as the SR difference between the first and last scenes 
in the sequence (\texttt{1$\sim$A} vs. \texttt{1$\sim$E}). A larger Drop indicates more severe catastrophic forgetting, 
providing a direct measure of knowledge retention across continual training:
\[
\text{Drop} = \text{SR}_{1\sim A} - \text{SR}_{1\sim E}.
\]

All continual methods share the same scene order, optimizer, batch size, learning rate schedule, and training epochs per scene. Replay capacity is aligned across replay-based methods (1280 samples). TinyER uses uniform random replay, while HyperDCM-VisFeat, HyperDCM-Euc, and HyperDCM-Hyp share identical capacity and replay ratio, differing only in representation or geometry. 
Joint NoMaD is trained on all scenes jointly and serves as an upper bound.

\subsection{Continual Learning Baselines}
We implement representative regularization-based methods 
(EWC~\cite{kirkpatrick2017overcoming}, SI~\cite{zenke2017continual}, LwF~\cite{li2017learning}) and replay-based 
TinyER~\cite{chaudhry2019tiny} under the same sequential protocol and memory budget. These methods cover regularization, distillation, and exemplar replay, and can be adapted to diffusion-policy training without changing the NoMaD-style backbone. We focus on these comparable baselines because more recent diffusion-specific continual-learning methods are less standardized for goal-conditioned navigation and often require additional generators, synthetic-data pipelines, or task-specific architectures.

EWC applies an online diagonal Fisher-based quadratic penalty estimated from replay samples. 
SI accumulates parameter importance during training and constrains important weights accordingly. 
LwF distills predicted noise outputs from the previous model via an MSE loss on replay samples, adapted to diffusion policies.

All regularization weights are selected via a small grid search under identical training schedules.
%



\subsection{Quantitative Results and Analysis}

Table~\ref{tab:performance_comparison} summarizes performance across five continual scene splits. 
As defined earlier, \texttt{1$\sim$X} denotes the model trained sequentially from Scene 1 to Scene X, evaluated using the checkpoint at the end of each stage.

Without replay, SR drops from 70.0\% at \texttt{1$\sim$A} to 26.7\% at \texttt{1$\sim$E} (Drop: 43.3\%), clearly indicating severe catastrophic forgetting. 
Regularization-based methods (EWC, SI, LwF) mitigate forgetting to some extent but remain limited, with Drop ranging from 37.4\% to 40.1\%.

Replay substantially improves retention. TinyER reduces the drop to 36.9\%, demonstrating the importance of revisiting past samples. 
Removing structural scene representation (HyperDCM-VisFeat) yields a Drop of 35.2\%, and centroid-aware selection in Euclidean space (HyperDCM-Euc) further lowers it to 34.8\%, highlighting the benefit of structured replay over uniform sampling.

HyperDCM-Hyp achieves the strongest retention with only 28.5\% Drop, while maintaining the highest SR across intermediate scenes. 
This consistent improvement suggests a clear progression: replay enhances stability, structural encoding improves sample diversity, and hyperbolic geometry further strengthens long-term retention in continual navigation. This progression across replay, structured replay, and hyperbolic replay confirms that geometry-aware memory organization provides complementary benefits beyond uniform sampling.

\begin{table}[htbp]
\centering
\caption{Performance comparison across different methods and scenes}
\resizebox{\columnwidth}{!}{
\begin{tabular}{lccccc|c}
\hline
\textbf{Method} & \textbf{1$\sim$A} & \textbf{1$\sim$B} & \textbf{1$\sim$C} & \textbf{1$\sim$D} & \textbf{1$\sim$E} & \textbf{Drop $\downarrow$} \\
\hline
\multicolumn{7}{c}{\textit{Upper Bound}} \\
Joint NoMaD            & 66.5 & 62.7 & 58.1 & 60.5 & 60.2 & 6.3 \\
\hline
\multicolumn{7}{c}{\textit{Continual Learning Methods}} \\
NoMaD-CL (No Replay)   & 70.0 & 40.6 & 34.4 & 25.2 & 26.7 & 43.3 \\
NoMaD-CL + EWC  & 67.4 & 42.3 & 35.7 & 30.2 & 27.3 & 40.1 \\
NoMaD-CL + SI   & 66.2 & 44.7 & 36.3 & 29.4 & 27.6 & 38.6 \\
NoMaD-CL + LwF   & 66.5 & 46.1 & 38.5 & 33.9 & 29.1 & 37.4 \\
NoMaD-CL + TinyER      & 65.5 & 45.5 & 37.6 & 31.5 & 28.6 & 36.9 \\
\hline
\multicolumn{7}{c}{\textit{Our Framework \& Ablations}} \\
HyperDCM-VisFeat (w/o Scene Graph) & 65.7 & 48.6 & 40.7 & 37.5 & 30.5 & 35.2 \\
HyperDCM-Euc (L2, Euclidean) & 66.7 & 48.2 & 41.1 & 36.0 & 31.9 & 34.8 \\
HyperDCM-Hyp (Poincaré, Ours)    & 70.0 & 50.8 & 55.2 & 45.2 & 41.5 & \textbf{28.5} \\
\hline
\end{tabular}
}
\label{tab:performance_comparison}
\end{table}

\begin{figure}[h]
    \centering
    \includegraphics[width=1\linewidth]{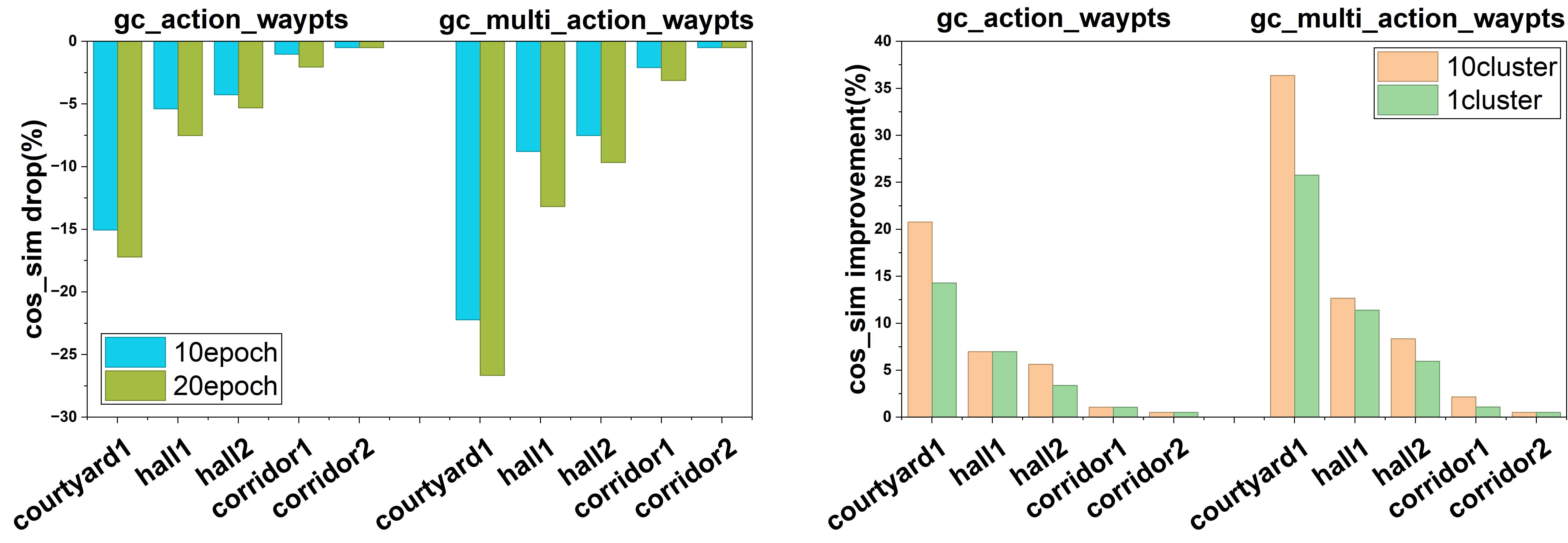}
    \vspace{-15pt}
    \caption{Trajectory-level forgetting analysis on goal-conditioned (GC) metrics. 
Left: Cosine similarity drop (\%) under 10 vs. 20 training epochs without replay. 
Longer training significantly amplifies catastrophic forgetting. 
Right: Cosine similarity improvement (\%) with structured memory (10 clusters) versus a single-buffer memory (1 cluster). 
Memory clustering consistently mitigates forgetting across continual scenes.}
    \label{fig:other_metric_compare}
\end{figure}

Beyond SR, we analyze trajectory-level GC cosine similarity metrics to examine forgetting at finer granularity. As shown in Fig.~\ref{fig:other_metric_compare}, increasing training epochs without replay amplifies the cosine similarity drop, indicating more severe forgetting. When memory size is fixed, clustering-based memory consistently outperforms a single buffer, confirming that structural organization improves retention.

\subsection{Online Metrics in Simulation}

Table II reports online navigation performance on five MatterPort3D~\cite{chang2017matterport3d} scenes 
used in the sequential continual learning protocol in the Habitat simulator~\cite{savva2019habitat}.
In addition to Success Rate (SR), we report SPL (Success weighted by Path Length) 
to measure trajectory efficiency, and collision frequency to quantify safety. 

Joint NoMaD serves as an upper bound via joint retraining. 
Compared to replay-free training, memory-based methods significantly improve 
SR and SPL while reducing collision frequency.
In particular, HyperDCM-Hyp achieves the best performance among continual 
learning methods, improving SR from 39.2\% to 56.0\% and SPL from 38.5\% 
to 48.3\%, while reducing collision frequency from 0.37 to 0.17 collisions per meter. Qualitative results are further illustrated in the supplementary video.

\begin{table}[htbp]
\centering
\caption{Online Evaluation in Habitat}
\resizebox{\columnwidth}{!}{
\begin{tabular}{lccc}
\hline
\textbf{Method} & \textbf{SR $\uparrow$} & \textbf{SPL $\uparrow$} & \textbf{Collision $\downarrow$} \\
\hline
\multicolumn{4}{c}{\textit{Upper Bound}} \\
Joint NoMaD & 63.1 & 52.7 & 0.13 \\
\hline
\multicolumn{4}{c}{\textit{Continual Learning Methods}} \\
NoMaD-CL (No Replay) & 39.2 & 38.5 & 0.37 \\
NoMaD-CL + TinyER & 41.7 & 40.2 & 0.29 \\
HyperDCM-VisFeat (w/o Scene Graph) & 45.0 & 42.0 & 0.24 \\
HyperDCM-Euc (L2, Euclidean) & 49.8 & 46.0 & 0.22 \\
HyperDCM-Hyp (Poincar\'e, Ours) & \textbf{56.0} & \textbf{48.3} & \textbf{0.17} \\
\hline
\end{tabular}
}
\label{tab:habitat_metrics}
\end{table}

\subsection{Ablation Study of HyperDCM}

To disentangle the contribution of structural scene representation and hyperbolic geometry, we evaluate two variants of our framework.

HyperDCM-VisFeat (w/o Scene Graph) removes the scene graph encoder and directly embeds EfficientNet visual features into hyperbolic space for memory replay, thereby isolating the effect of structural scene representation. 

HyperDCM-Euc (L2, Euclidean) retains the centroid-based replay mechanism but replaces hyperbolic distance with Euclidean L2 distance, isolating the geometric effect.

As shown in Table~\ref{tab:performance_comparison}, removing structural representation or replacing hyperbolic geometry both degrade retention performance, indicating that hierarchical scene encoding and hyperbolic geometry jointly contribute to continual stability. The gap between HyperDCM-Euc and HyperDCM-Hyp suggests that hyperbolic curvature helps separate structurally similar but task-specific memories, reducing replay interference compared with Euclidean memory.






\subsection{Qualitative Results}
Fig.~\ref{fig:eval_show} visualizes predicted waypoint trajectories across four representative scenes. The continual NoMaD baseline exhibits severe divergence on earlier scenes due to forgetting, while replay-based methods recover more faithful predictions. Notably, HyperDCM-Hyp preserves both path smoothness and goal alignment across all scenes, confirming that structure-sensitive replay not only maintains performance numerically but also yields more interpretable navigation behavior.

\subsection{Training Efficiency and Adaptation Performance}

We compare HyperDCM with joint retraining, which revisits 
all accumulated data at each stage. Training cost is measured 
as diffusion policy optimization time (GPU-hours). Scene graph 
extraction is performed once per scene as offline preprocessing 
and reused across stages, and the reported GPU-hours reflect 
only diffusion policy training. As shown in Fig.~6, joint retraining incurs rapidly increasing 
training cost as scenes accumulate, whereas HyperDCM 
maintains nearly constant per-stage training cost.

We report the success rate on the latest scene to measure 
adaptation performance. Although joint retraining achieves 
stronger overall performance, HyperDCM maintains competitive adaptation performance while requiring substantially lower diffusion retraining cost.

\begin{figure}[t]
    \centering
    \includegraphics[width=1\linewidth]{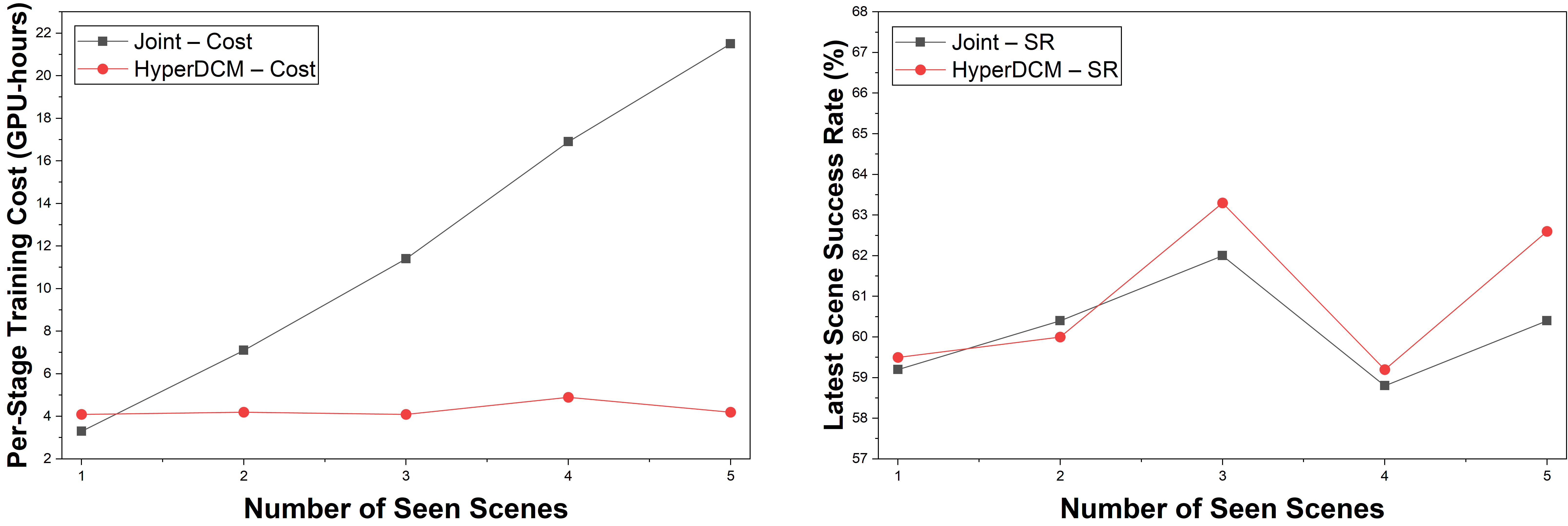}
    \vspace{-10pt}
    \caption{Comparison between joint retraining and HyperDCM under sequential scene training. 
Left: Per-stage training cost (GPU-hours). Joint retraining exhibits approximately linear growth as the number of accumulated scenes increases, while HyperDCM remains nearly constant. 
Right: Success rate on the latest scene at each stage, reflecting adaptation performance to newly encountered environments.}
    \label{fig:efficiency_tradeoff}
\end{figure}

\section{CONCLUSION}
In this work, we introduced HyperDCM, a structure-aware hyperbolic memory mechanism for continual diffusion-based navigation.
The proposed dynamic clustering and structure-sensitive memory update strategy enhances the representativeness of replayed samples, alleviating catastrophic forgetting in multi-scene navigation tasks. 
Experimental results on indoor and outdoor datasets show that HyperDCM retains knowledge more effectively and achieves better generalization than NoMaD when applied in continual learning settings. 
Our work highlights the importance of continual learning for diffusion-based navigation. Beyond navigation, our work demonstrates the significant potential of combining structural world modeling with hyperbolic geometry to address catastrophic forgetting in robotic learning. 
In future work, we will explore adaptive memory allocation strategies and integration with model-generated replay samples further to improve scalability and adaptability in open-world navigation scenarios.




\bibliographystyle{IEEEtran}
\bibliography{references}

\end{document}